\documentclass[runningheads]{llncs}
\usepackage{subfigure}
\usepackage{graphicx}
\usepackage{float}
\usepackage{amsmath}
\usepackage{amssymb}
\usepackage{xcolor}
\usepackage[hidelinks]{hyperref}
\usepackage{xspace}
\usepackage{booktabs}
\usepackage{multirow}
\usepackage[ruled,linesnumbered,algo2e]{algorithm2e}
\usepackage{bbding}
\usepackage[marginal]{footmisc}

\usepackage[misc]{ifsym}
\newcommand{\ourmethod}{\texttt{RawNP}\xspace}
\begin{document}
\title{Towards Few-shot Inductive Link Prediction on Knowledge Graphs: A Relational Anonymous Walk-guided Neural Process Approach}
\titlerunning{RawNP: Few-shot Inductive Link Prediction on Knowledge Graphs}
\author{
Zicheng Zhao \inst{1,2}* \and
Linhao Luo \inst{3}* \and
Shirui Pan \inst{4} \and
Quoc Viet Hung Nguyen \inst{4} \and
Chen Gong \inst{1,5} \Letter
}
\authorrunning{Z. Zhao, L. Luo et al.}
\institute{School of Computer Science and Engineering, Nanjing University of Science and Technology \and
Jiangsu Key Laboratory of Image and Video Understanding for Social Security \and
Department of Data Science and AI, Monash University \and
School of Information and Communication Technology, Griffith University \and
Key Laboratory of Intelligent Perception and Systems for High-Dimensional Information of Ministry of Education \\
\email{chen.gong@njust.edu.cn}
}

\toctitle{Towards Few-shot Inductive Link Prediction on Knowledge Graphs: A Relational Anonymous Walk-guided Neural Process Approach}
\tocauthor{Zicheng~Zhao,~Linhao~Luo,~Shirui~Pan,~Quoc~Viet~Hung~Nguyen,~Chen~Gong}

\maketitle              
\footnote{* Equal contribution.}

\begin{abstract}
Few-shot inductive link prediction on knowledge graphs (KGs) aims to predict missing links for unseen entities with few-shot links observed. Previous methods are limited to transductive scenarios, where entities exist in the knowledge graphs, so they are unable to handle unseen entities. Therefore, recent inductive methods utilize the sub-graphs around unseen entities to obtain the semantics and predict links inductively. However, in the few-shot setting, the sub-graphs are often sparse and cannot provide meaningful inductive patterns. In this paper, we propose a novel \textbf{r}elational \textbf{a}nonymous \textbf{w}alk-guided \textbf{n}eural \textbf{p}rocess for few-shot inductive link prediction on knowledge graphs, denoted as \ourmethod. Specifically, we develop a neural process-based method to model a flexible distribution over link prediction functions. This enables the model to quickly adapt to new entities and estimate the uncertainty when making predictions. To capture general inductive patterns, we present a relational anonymous walk to extract a series of relational motifs from few-shot observations. These motifs reveal the distinctive semantic patterns on KGs that support inductive predictions. Extensive experiments on typical benchmark datasets demonstrate that our model derives new state-of-the-art performance.

\keywords{Knowledge graphs \and Few-shot learning \and Link prediction \and Neural process.}
\end{abstract}
\section{Introduction} 
Knowledge graphs (KGs) are structured representations of human knowledge, where each link represents the fact in the format of a triple ($head~entity$, $relation$, $tail~entity$). Recently, KGs have been widely used in various applications, such as web search \cite{luo2023gsim}, community detection \cite{luo2021detecting} and recommender systems \cite{wang2019multi}. However, the incompleteness of KGs \cite{zhang2023multiform} largely impairs their applications. 
Therefore, many methods have been proposed to complete KGs by predicting the missing links and they achieved impressive performances \cite{bordes2013translating,menon2011link}.

\begin{figure}[t]
    \centering
    \includegraphics[width=.8\linewidth]{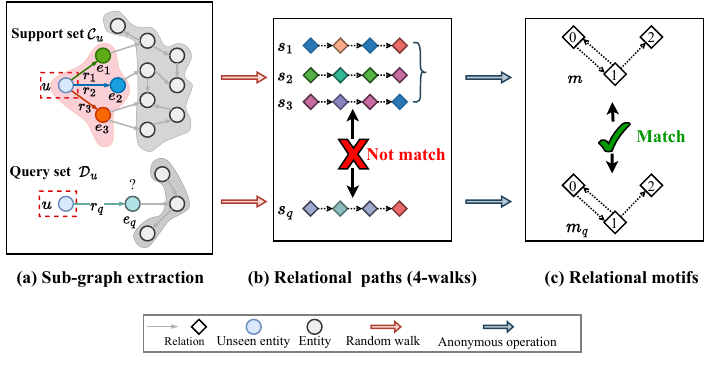}
    \caption{An illustration of few-shot inductive link prediction on knowledge graphs.}
    \label{fig:introduction}
\end{figure}

Despite the success, these traditional methods are often transductive, assuming that all entities are seen during training. However, real-world KGs dynamically evolve over time, with numerous unseen entities emerging every day \cite{luo2023graph,shi2018open}. In this case, transductive methods can hardly model the unseen entities, resulting in an unsatisfactory performance for inductive link prediction. Moreover, unseen entities usually have few links upon their arrival \cite{wanglearning}, thus providing insufficient information to characterize themselves. Therefore, few-shot inductive link prediction on KGs has recently attracted increasing attention \cite{baek2020learning,chen2022meta,wang2022exploring}. As shown in Fig.~\ref{fig:introduction}(a), given an unseen node $u$ and its support set $\mathcal{C}_u$ with three observed links ($r_1$, $r_2$, $r_3$), few-shot inductive link prediction aims to predict possible links $r_q$ with other entities $e_q$ in the query set $\mathcal{D}_u$.

Inspired by graph neural networks (GNNs) \cite{wan2021contrastive}, recent studies utilize a sub-graph around the unseen entity to predict links inductively \cite{mai2021communicative,teru2020inductive,zheng2022subgraph}. The major motivation behind these methods is that they try to capture the semantic patterns from the graph topology that are agnostic to the target entity. The semantic patterns on KGs are usually reflected as relational paths \cite{wang2021relational,xusubgraph}, and each of them is a sequence of relations connecting the entities, as shown in Fig.~\ref{fig:introduction}(b). Given the observations, an ideal pattern for inductive reasoning should be distinctive and can be matched during the inference. However, in the few-shot setting, the sub-graphs are often sparse, making semantic features captured by relational paths \textit{not general enough and meaningful for inductive link prediction} (\textbf{Limitation 1}). For example, as shown in Fig.~\ref{fig:introduction}(b), the three relational paths (i.e., $s_1\sim s_3$) extracted from the support set are distinct, where different colors denote different relations. They cannot provide any inductive and distinctive patterns that can be matched by the relational path extracted from the query set. Therefore, the patterns represented by relational paths cannot be used for supporting the inductive link prediction.

Recently, several meta-learning-based methods have been proposed to tackle the problem of few-shot learning. Meta-learning-based methods quickly adapt to a new entity by updating the model parameters with a few examples \cite{brazdil2022advances,zheng2022subgraph}. However, due to the limitation of data, the meta-learning-based methods often suffer from overfitting \cite{dong2020mamo} and out-of-distribution problems \cite{huangfew}. Meanwhile, they \textit{fail to quantify the uncertainty in the predictions} (\textbf{Limitation 2}), which is essential for generating reliable predictions under few-shot scenarios \cite{xiao2021hmnet,zhanguncertainty}.

To address the aforementioned limitations, we propose a \textbf{r}elational \textbf{a}nonymous \textbf{w}alk-guided \textbf{n}eural \textbf{p}rocess approach (\ourmethod) for few-shot inductive link prediction on knowledge graphs. Specifically, we develop a framework based on neural process (NP) \cite{garnelo2018neural} to address the challenges mentioned above in few-shot learning. Unlike previous few-shot methods (e.g., meta-learning), NP is based on the stochastic process that models the distribution over the functions conditioned on limited data. Given a few links, we can readily obtain a prediction function from the distribution that is specialized for the unseen entity. By modeling the distribution, \ourmethod can also estimate the uncertainty of its prediction and generate more reliable results (\textbf{addressing Limitation 2}). To capture the representative patterns, we propose a novel relational anonymous walk to extract a series of relational motifs (\textbf{addressing Limitation 1}). As shown in Fig.~\ref{fig:introduction}(c), the three relational paths (i.e., $s_1\sim s_3$) in Fig.~\ref{fig:introduction}(b) can be represented by one relational motif $m$, which can be used to guide the inductive predictions by matching with the motif $m_q$ in the query set. The main contributions of our work are summarized as follows:
\begin{itemize}
    \item We propose a novel neural process approach for few-shot inductive link prediction on knowledge graphs. To the best of our knowledge, this is the first work of developing a neural process framework to solve this problem.
    \item We propose a novel relational anonymous walk to extract a series of relational motifs. The patterns revealed from these motifs are more general and distinctive than the previous methods for inductive link prediction.
    \item We conduct extensive experiments on typical public datasets. Experimental results show that \ourmethod outperforms existing baseline methods, which proves the superiority of our method.
\end{itemize}
\section{Related Work}

\textbf{Link Prediction on Knowledge Graphs.} 
Link prediction on knowledge graphs is an important task to complete the missing facts. Previous methods mainly focus on the transductive setting, where all entities are seen during training \cite{bordes2013translating,wan2021reasoning,yang2014embedding}. Inspired by the inductive ability of GNNs \cite{wan2022multi}, several methods adopt the graph structure to predict links inductively \cite{chen2021topology,liu2021indigo}. For example, GraIL \cite{teru2020inductive} extracts the enclosing sub-graph of a given triple to capture the topological structure. CoMPILE \cite{mai2021communicative} generates inductive representations by modeling the relations in sub-graphs. To better consider the semantics in knowledge graphs, SNRI \cite{xusubgraph} adopts relational paths within a sub-graph to provide inductive features. However, the features captured by these methods are not general enough to provide inductive bias for unseen entities, especially in the few-shot setting. Meanwhile, there are several works \cite{jin2022neural,wang2021inductive} that apply anonymous random walk on temporal graphs to extract temporal network motifs, thus keeping their methods fully inductive. However, they focus on node anonymization and cannot handle complex relations in  knowledge graphs.

Several meta-learning-based methods have been proposed for few-shot link prediction. MetaR \cite{chen2019meta} adapts to unseen relations by a relation-meta learner and updates the parameter by using the meta-learning framework. Meta-iKG \cite{zheng2022subgraph} utilizes local sub-graphs to transfer sub-graph-specific information and rapidly learn transferable patterns via meta-learning. However, the meta-learning-based methods are sensitive to the quality of given few-shot data and unable to estimate the uncertainty of the model. GEN \cite{baek2020learning} meta-learns the unseen node embedding for inductive inference and proposes a stochastic embedding layer to model the uncertainty in the link prediction, which achieves state-of-the-art performance among all baseline models.

\vspace{2mm}
\noindent\textbf{Neural Process.} 
Neural process (NP) \cite{garnelo2018neural}, a new family of methods, opens up a new door to dealing with limited data in machine learning \cite{wan2021contrastive2021}. Based on the stochastic process, NP enables to model the distribution over functions given limited observations and provides an uncertainty measure to the predictions. An increasing number of researches focus on improving the expressiveness of the vanilla NP model. For instance, Attentive Neural Process (ANP) \cite{kim2019attentive} leverages the self-attention mechanism to better capture the dependencies and model the distribution. Sequential Neural Process (SNP) \cite{singh2019sequential} introduces a recurrent neural network (RNN) to capture temporal correlation for better generalization. NP has already been applied in many tasks to address the challenge of data limitation, such as recommender systems \cite{lin2021task}, node classification \cite{cangea2022message}, and link prediction \cite{liang2022neural,luo2022graph}. This also demonstrates the great potential of NP in other machine learning areas. Recently, NP-FKGC \cite{luo2023normalizing} applies normalizing flow-based NP to predict the missing facts for few-shot relations. To the best of our knowledge, this is the first work to apply the neural process to the few-shot inductive link prediction on knowledge graphs.

\section{Preliminary and Problem Definition}
\subsection{Neural Process} 
NP \cite{garnelo2018neural} marries the benefits of the stochastic process and neural networks to model the distribution over functions $f: X \rightarrow Y$ with limited data, where $X$ and $Y$ are feature space and label space, respectively. Specifically, the function $f$ is assumed to be parameterized by a high-dimensional random vector $z$, whose distribution $P(z|\mathcal{C})$ is conditioned on the \textit{context data} $\mathcal{C}=\left\{\left(x_\mathcal{C}, y_\mathcal{C}\right)\right\}$ with $x$ and $y$ denoting feature and label of a data point accordingly. The $P(z|\mathcal{C})$ is empirically defined as a Gaussian distribution, which is modeled by an \textit{encoder} using the context data. By sampling a $z$ from the distribution, NP can easily obtain the function for a new prediction task. The prediction likelihood over the \textit{target data} $\mathcal{D}=\left\{\left(x_\mathcal{D}, y_\mathcal{D}\right)\right\}$ is calculated as

\begin{equation}
\setlength\abovedisplayskip{1pt}
\setlength\belowdisplayskip{1pt}
\label{equ:equ1}
P\left(y_\mathcal{D}|x_\mathcal{D}, \mathcal{C}\right)=\int_z P\left(y_\mathcal{D}|x_\mathcal{D}, z\right) P(z|\mathcal{C})dz,
\end{equation}
where $P\left(y_\mathcal{D}|x_\mathcal{D}, z\right)$ is modeled by a \textit{decoder} network. Since the actual distribution of $z$ is intractable, the training of NP can be achieved by amortized variational inference \cite{kingma2013auto}. The objective expressed by Eq.~\eqref{equ:equ1} can be optimized by maximizing the \textbf{e}vidence \textbf{l}ower \textbf{bo}und (ELBO), which is formulated as 
\begin{equation}
    \label{equ:equ2}
    \begin{split}
    \log P\left(y_\mathcal{D}|x_\mathcal{D}, \mathcal{C}\right) \geq \mathbb{E}_{Q_\psi}(z|\mathcal{C}, \mathcal{D})\left[\log P_\phi\left(y_\mathcal{D}|x_\mathcal{D}, z\right)\right]
    -KL\left(Q_\psi(z|\mathcal{C}, \mathcal{D}) \| P_\theta(z|\mathcal{C})\right),
    \end{split}
\end{equation}
where $\theta$ and $\phi$ denote the parameters of encoder and decoder, respectively, and $Q_{\psi}(z|\mathcal{C}, \mathcal{D})$ denotes the variational posterior of the latent variable $z$, approximated by another neural network with parameters $\psi$.

\subsection{Problem Definition}
A  KG can be represented by a set of triples $\mathcal{G}=\{(h, r, t) \subseteq \mathcal{E} \times \mathcal{R} \times \mathcal{E}\}$, where $\mathcal{E}$ and $\mathcal{R}$ denote the set of existing entities and relations in KG respectively, $h, t \in \mathcal{E}$ denote the head and tail entities and $r \in \mathcal{R}$ denotes the specific relations between the entities. The few-shot inductive link prediction on KGs can be formulated as follows:
\begin{definition}
\textbf{Few-shot inductive link prediction on knowledge graphs.} Given a knowledge graph $\mathcal{G}$ and an unseen entity set $\mathcal{\widetilde{E}}$, where $\mathcal{E} \cap \mathcal{\widetilde{E}}=\emptyset$, we assume that each unseen entity $u\in \mathcal{\widetilde{E}}$ is associated with a $K$-shot \textit{support set} $\left\{\left(u, r_i, e_i\right)\right\}_{i=1}^K$, where $e_i \in\mathcal{E}\cup\mathcal{\widetilde{E}}$. For an unseen entity $u$, our task is to obtain a function $f_u$ that predicts the other entity $e_q$ for each query $q=\left( u, r_q, ? \right)$ in the \textit{query set} $\{\left( u, r_q, ? \right)\}$, where $e_q\in\mathcal{E}\cup\mathcal{\widetilde{E}}$ and $r_q$ is the given query relation.
\end{definition}

In our paper, we propose a neural process-based framework for this task. For each unseen entity $u$, we treat its support set as the context data $\mathcal{C}_u=\left\{\left(u, r_i, e_i\right)\right\}_{i=1}^K$ and the query set as the target data $\mathcal{D}_u= \{\left( u, r_q, ? \right)\}$.

\section{Approach}
\begin{figure}[t]
    \centering
    \includegraphics[width=\linewidth]{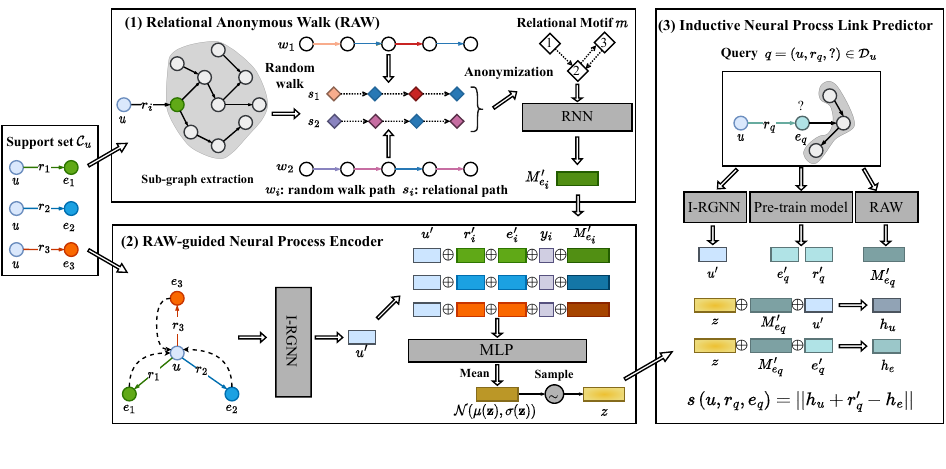}
    \caption{The framework of our proposed model \ourmethod for few-shot inductive link prediction on knowledge graphs.}
    \label{fig:architecture}
 \end{figure}

In this section, we present our proposed model \ourmethod, which consists of three major components: (1) a relational anonymous walk (RAW) to generate a series of relational motifs for each entity and excavate distinctive semantic patterns; (2) a RAW-guided neural process encoder to model the joint distribution over link prediction functions on knowledge graphs and simultaneously estimate the uncertainty for predictions; (3) an inductive neural process link predictor to infer the inductive links given an unseen entity and its associated relation. The overall framework of our proposed model is illustrated in Fig.~\ref{fig:architecture}. 

\subsection{Relational Anonymous Walk}
\label{Raw}
The relational anonymous walk (RAW) is designed to capture the distinctive semantic patterns on KGs, which better reveal the inductive identity and facilitate the link prediction. Previous methods capture the semantic patterns in the sub-graphs around entities by using relational path \cite{wang2021relational,xusubgraph}, which is a sequence of relations connecting the entities. Specifically, given a raw path in the knowledge graphs: $w=e_0\xrightarrow{r_1}e_1\xrightarrow{r_2}\ldots\xrightarrow{r_l}e_{l}$, the corresponding relational paths $s$ is the sequence of relations in the given path, i.e., $s=\{r_1,r_2,\ldots,r_l\}$. However, the relational path is not general enough, as the combinations of relations explode in KGs, making the patterns captured by relational paths not distinguishable. To address this issue, we propose a relational anonymous walk to extract the distinctive semantic patterns in the form of relational motifs. The process of RAW is shown in Algorithm~\ref{alg:raw}. 


For each entity $e$ in the triple $(u,r,e)$ of the $K$-shot support set, we first perform random walk \cite{perozzi2014deepwalk} to sample a few $l$-step paths $\{w_i=e\xrightarrow{r_1}e_1\xrightarrow{r_2}\ldots\xrightarrow{r_l}e_{l}\}_{i=1}^L$ starting from $e$, where $L$ denotes the number of walks. Then, we could obtain the corresponding relational paths $\{s_i=\{r_1,r_2,\ldots,r_l\}\}_{i=1}^L$, where relations could be repeated in $s_i$. Later, we apply an anonymization operation $A(\cdot)$ to each $s_i$ by replacing the actual relations with their first positions in $s_i$. This can be formulated as
\begin{gather}
    m_i = A(s_i)=\{I(r_1), I(r_2),\ldots, I(r_l)\}, \\
    I(r_j) = \min~pos(r_j, s_i),
\end{gather}  
where $pos(r_j,s_i)\in [1, l]$ denotes the positions of $r_j$ in $s_i$.
The anonymization operation $A(\cdot)$ removes the relation identities and maps the relational paths into a general semantic pattern defined as a \textit{relational motif} $m_i$. For example, as shown in the bottom of Fig.~\ref{fig:architecture}, the two distinct relational paths $s_1$ and $s_2$ can be anonymized to the same relational motif structure $m$. In this way, by checking the set of relational motifs $\mathcal{M}_{e}$, we can find the distinctive features for inductive link predictions. 

To obtain the representation of patterns, we first encode each motif $m_i \in \mathcal{M}_e$ by using a recurrent neural network (RNN) and aggregate them with a mean pooling, which is formulated as
\begin{gather}
    m_i' = \text{RNN}(\{f_{enc}\big(I(r_j)\big)|I(r_j)\in m_i\}),\\
    M'_{e} = \frac{1}{|\mathcal{M}_{e}|}\sum_{m_i \in \mathcal{M}_{e}} m'_i,
\end{gather}
where $f_{enc}$ is a multi-layer perceptron (MLP) mapping function.
\begin{algorithm2e}[t]
    \caption{Relational anonymous walk (RAW)}\label{alg:raw}
    \KwIn {Knowledge graph $\mathcal{G}$; unseen entity $u$; support triple $(u, r, e)\in \mathcal{C}_u$; walks number $L$; walks length $l$}
    \KwOut {Relational motifs set $\mathcal{M}_{e}$}
    Initialize $\mathcal{M}_{e}\leftarrow\emptyset$\;
    \For{i=1 to L}{
        Sample a $l$-step path $w_i$ starting from $e$ using random walk\;
        Obtain the corresponding relational path $s_i$\;
        Apply anonymization operation $A(s_i)$ to extract the motif $m_i$\;
        Add $m_i$ to $\mathcal{M}_{e}$\;
    }
\end{algorithm2e}

\subsection{RAW-guided Neural Process Encoder}\label{sec:enc}
RAW-guided neural process encoder attempts to model the joint distribution over the link prediction functions based on the context data (support set). It first learns a low-dimension vector $c_i$ for each triple in the context data. Then, it aggregates them into a global representation $\mathbf{z}$, which defines the distribution as $\mathcal{N}(\mu(\mathbf{z}),\sigma(\mathbf{z}))$. By sampling a $z$ from the distribution, we can adaptively obtain the prediction function $f_u$.

For each triple $(u, r_i, e_i)$ in the support set, we first adopt RAW to obtain the pattern representation $M'_{e_i}$ to inject the inductive ability into $f_u$. Since the relational motifs ignore the identity information, we also obtain the representations of entity $e'_i$ and relation $r'_i$ from a pre-trained model, e.g., TransE \cite{bordes2013translating}. For the unseen node $u$, we adopt an inductive relational graph neural network (I-RGNN) to generate the representation by aggregating all the triples in its support set, which is formulated as
\begin{equation}
    u' = ReLU(\frac{1}{|\mathcal{C}_u|}\sum_{(u,r_i,e_i)\in\mathcal{C}_u} W_{r_i}r'_i + We'_i),
\end{equation}
where $W_{r_i}$ denotes a relation-specific weight matrix and $W$ is a weight matrix. Through aggregating from associated triples in the support set, I-RGNN enables the inductive generation of embeddings for unseen entities.

By incorporating the representations of $u'$, $r'_i$, $e'_i$, and $M'_{e_i}$, $c_i$ is generated as follows:
\begin{equation}
    c_i=\operatorname{MLP}\left(u^{\prime}\left\|r_i^{\prime}\|e_i^{\prime}\right\| y_i\|M'_{e_i}\right), y_i=\left\{\begin{array}{l}
    1,\left(u, r_i, e_i\right) \in \mathcal{C}_u \\
    0,\left(u, r_i, e_i\right) \in \mathcal{C}_u^{-}
    \end{array}\right.,
\end{equation}
where we sample a set of negative samples $\mathcal{C}_u^{-}$ by replacing the $e_i$ in $\mathcal{C}_u$ with the other entities randomly, and $y_i$ is an indicator vector. $\|$ represents concatenation operations.

Then, we aggregate all the latent representations $c_i \in \mathcal{C}_u \cup \mathcal{C}_u^{-}$ to obtain a global representation $\mathbf{z}$ and define the joint distribution over the link prediction functions. The aggregator function must satisfy the condition of \emph{permutation-invariant} \cite{garnelo2018neural,van1976stochastic}. Therefore, we select the mean pooling function, which can be formulated as
\begin{align}
    \mathbf{z}=\frac{1}{\left|\mathcal{C}_u \cup \mathcal{C}_u^{-}\right|} \sum_{c_i \in \mathcal{C}_u \cup \mathcal{C}_u^{-}} c_i.
\end{align}

The distribution $P(\mathbf{z}|\mathcal{C}_u)$ is empirically considered as a Gaussian distribution $\mathcal{N}(\mu(\mathbf{z}), \sigma(\mathbf{z}))$ parameterized by $\mathbf{z}$ \cite{garnelo2018conditional,kim2019attentive}, in which the mean $\mu(\mathbf{z})$ and variance $\sigma(\mathbf{z})$ are modeled by two neural networks:
\begin{gather}
    h_z=\emph{ReLU}(\operatorname{MLP}(\mathbf{z})), \\
    \mu(\mathbf{z})=\operatorname{MLP}(h_z), \\
    \sigma(\mathbf{z})=0.1+0.9 * \emph{Sigmoid}(\operatorname{MLP}(h_z)).
\end{gather}

Noticeably, $\mathcal{N}(\mu(\mathbf{z}), \sigma(\mathbf{z}))$ not only defines the distribution over functions, but also estimates the uncertainty of the model. When the support set is limited, the encoder could generate a distribution with a larger variance, which indicates that the model is more uncertain to its predictions. We detailly analyze the uncertainty captured by \ourmethod in Section~\ref{sec:uncertainty}.

\subsection{Inductive Neural Process Link Predictor}
The inductive neural process link predictor serves the decoder to realize the $f_u$ modeled by $\mathcal{N}(\mu(\mathbf{z}), \sigma(\mathbf{z}))$. Given a query $q=\left(u, r_q, ?\right)$, $f_u$ tries to predict the possible entity $e_q$. The details are as follows:

In the predictor, we obtain the representations of entities and relations (i.e., $u',r_q',e'_q$) with the same process in Section~\ref{sec:enc} and combine them with a sampled $z$ by following the paradigm of neural process, which is calculated as
\begin{equation}
    \operatorname{Sample}~z\sim \mathcal{N}(\mu(\mathbf{z}),\sigma(\mathbf{z})),
\end{equation}
where each sample of $z$ is regarded as a realization of the function from corresponding stochastic process. 

Then, we use two independent MLPs to map $z$ into the space of entities and project $u^{\prime}$ and $e_q^{\prime}$ into the hyper-planes defined by $z$ via using an element-wise addition, which can be formulated as
\begin{gather}
    u^{\prime}_z = u^{\prime} + \operatorname{MLP}_{u}^z(z), e^{\prime}_z = e_q^{\prime} + \operatorname{MLP}_{e}^z(z).
\end{gather}

For inductive prediction, we also obtain the relational motif representation $M'_{e_q}$ produced by the relational anonymous walk introduced in Section~\ref{Raw}. Similarly, we inject this representation by another two MLPs, which are formulated as
\begin{align}
    h_u = u^{\prime}_z + \operatorname{MLP}_{u}^{M} (M'_{e_q}), h_e = e^{\prime}_z + \operatorname{MLP}_{e}^M (M'_{e_q}).
\end{align}

Finally, we use a score function to measure the plausibility of triples, which is formulated as
\begin{align}
    s\left(u, r_q, e_q\right) = || h_u + r_q' - h_e||.
\end{align}

\subsection{Optimization and Inference}
\textbf{Optimization.} Given an unseen entity $u$ and its support set $\mathcal{C}_u$, our objective is to infer the distribution $P(z|\mathcal{C}_u)$ from the context data that minimizes the prediction loss on the target data $\log P\left(e_q|u, r_q, \mathcal{C}_u\right)$. The optimization can be achieved by maximizing the \textbf{e}vidence \textbf{l}ower \textbf{bo}und (ELBO), as derived:
\begin{align}
    \small
    & \log P\left(e_q|u, r_q, \mathcal{C}_u\right)=\int_z Q(z) \log \frac{P\left(e_q,z|u, r_q,\mathcal{C}_u\right)}{P\left(z|\mathcal{C}_u\right)}, \\
    &=\int_z Q(z) \log \frac{P\left(e_q,z|u, r_q, \mathcal{C}_u\right)}{Q(z)}+KL\left(Q(z) \| P\left(z|\mathcal{C}_u\right)\right), \\ 
    & \geq \int_z Q(z) \log \frac{P\left(e_q,z|u, r_q, \mathcal{C}_u\right)}{Q(z)}, \\ 
    &=\mathbb{E}_{Q(z)} \log \frac{P\left(e_q,z|u, r_q, \mathcal{C}_u\right)}{Q(z)}, \\ 
    &=\mathbb{E}_{Q(z)}\left[\log P\left(e_q|u, r_q, z\right)+\log \frac{P\left(z|\mathcal{C}_u\right)}{Q(z)}\right], \\ 
    &=\mathbb{E}_{Q(z)}\left[\log P\left(e_q|u, r_q, z\right)\right]-K L\left(Q(z) \| P\left(z|\mathcal{C}_u\right)\right),\label{equ:elbo_loss}
\end{align}
where $Q(z)$ represents the true posterior distribution of $z$, which is intractable. To address this problem, we approximate it with $Q\left(z|\mathcal{C}_u, \mathcal{D}_u\right)$ calculated by the encoder during training. The detailed derivation of Eq.~\eqref{equ:elbo_loss} can be found in the \emph{Appendix}.

We introduce the \emph{reparamterization trick} for sampling $z$ to support gradient propagation, and then we estimate the expectation $\mathbb{E}_{Q(z)}\left[\log P\left(e_q|u, r_q, z\right)\right]$ via the Monte-Carlo sampling as follows:
\begin{align}
    \small
    \mathbb{E}_{Q(z)}\left[\log P\left(e_q|u, r_q, z\right)\right] & \simeq \frac{1}{T} \sum_{t=1}^T \log P\left(e_q|u, r_q, z^{(t)}\right), \\ 
    z^{(t)}=\mu(\mathbf{z})+\sigma(\mathbf{z}) \epsilon^{(t)}, & \text { with } \epsilon^{(t)} \sim \mathcal{N}(0,1).
\end{align}

The likelihood term $\log P\left(e_q|u, r_q, z\right)$ is calculated by a widely-used margin ranking loss as follows:
\begin{align}
    \log P\left(e_q|u, r_q, z\right) = -\sum_{q,q^-}max\left(0,\gamma +s\left(q^-\right)-s\left(q\right)\right),
\end{align}
where $\gamma$ denotes a margin hyper-parameter, and $q=(u,r_q,e_q)$ denotes the ground truth triples, and $q^-$ denotes the negative triples by randomly corrupting $e_q$. By maximizing the likelihood, we aim to rank the scores of positive triples higher than all other negative triples.

\textbf{Inference.} In the inference stage, given an unseen entity $u$, we generate latent distribution $P\left(z|\mathcal{C}_u\right)$ by using its support set $\mathcal{C}_u$. Then, we feed the sampled $z$ together with the embeddings of unseen entity $u$ and its query relation $r_q$ to the decoder and predict the possible entity $e_q$ for the target set $\mathcal{D}_u$. The algorithms of the training and testing process can be found in the \emph{Appendix}.
\section{Experiment}

\subsection{Datasets and Evaluation}
We conduct our experiments on two benchmark datasets: FB15k-237 \cite{toutanova2015representing} and NELL-995 \cite{xiong2017deeppath}. To support the inductive setting, we randomly filter a few entities out of KGs as unseen entities. For FB15k-237 dataset, we first select 5000 entities whose related triples are between 10 to 100 and split them into 2,500/1,000/1,500 for training/validation/test. For NELL-995 dataset, we choose 3000 entities whose associated triples are between 7 to 100 and split them into 1,500/600/900 for training/validation/test. The splits are following the same settings in GEN \cite{baek2020learning} and the statistics of two datasets can be found in \emph{Appendix}.

In the evaluation stage, for a query triple $(u, r_q, ?)$, we construct the candidate set by using all the possible entities in the KG. We obtain the rank of the correct triples and report the results using the mean reciprocal rank (MRR) and the Top-N hit ratio (Hits@N). The $N$ is set to 1, 3, and 10 to directly compare with the existing methods.

\subsection{Baseline Models}
We select a series of following baseline models for comparison, which can be divided into three categories: (1) \textbf{Traditional KGC methods}, including TransE \cite{bordes2013translating}, DistMult \cite{yang2014embedding}, ComplEx \cite{trouillon2016complex}, RotatE \cite{sun2019rotate}; (2) \textbf{GNN-based methods}, including R-GCN \cite{schlichtkrull2018modeling}, MEAN \cite{hamaguchi2017knowledge}, LAN \cite{wang2019logic}; (3) \textbf{Few-shot inductive methods}, including GMatching \cite{xiong2018one}, MetaR \cite{chen2019meta}, FSRL \cite{zhang2020few}, GEN \cite{baek2020learning}. Specifically, there are two versions of the GEN model: I-GEN, which does not consider relations between unsees entities, and T-GEN which remedies the defect. More details can be found in the \emph{Appendix}. To avoid the re-implementation bias, we directly use the existing SOTA results reported by GEN \cite{baek2020learning} in experiments.

\subsection{Implementation Details}
We implement our model with PyTorch \cite{paszke2019pytorch} and PyG \cite{fey2019fast} package and conduct experiments on a single RTX 3090 GPU. The dimensions of entity and relation embedding are set to 100. The length of random walk  $l$ in the relational motifs extractor is set to 10, and the walk number $L$ is set to 5. We set the learning rate as $10^{-3}$, margin $\gamma$ as 1, dropout rate as 0.3, and negative sample size as 32 and 64 in FB15k-237 and NELL-995, respectively. We use Adam as the optimizer. 
We use the pre-trained model (e.g., TransE \cite{bordes2013translating}) to initialize the embeddings of entity and relation, which is fine-tuned during training. We set the embedding of unseen entities as the zero vector. 
Finally, the best model used for testing is selected according to the metric of MRR on the evaluation set. More detailed experiment settings can be found in the \emph{Appendix}. Code and appendix are available at \url{https://github.com/leapxcheng/RawNP}.

\subsection{Results and Analysis}

We present the results of 1-shot and 3-shot link prediction on FB15k-237 and NELL-995 in Table~\ref{tab:main-table}, where the best results are highlighted in bold. From the results, we can see that our \ourmethod achieves the best performance against all baseline models, demonstrating the superiority and effectiveness of our model. 

Traditional KGC methods get the worst results.
Because they cannot well represent the emerging unseen entity and barely works under the inductive setting. GNN-based methods achieve better performance as they consider the local structure of the knowledge graph. Specifically, LAN uses the attention mechanism to capture the semantics inherent in the knowledge graph, which achieves the best performance among GNN-based methods. However, with the limitation of the data (e.g., 1-shot), their performance drops quickly.
Few-shot methods focus on making predictions with limited data and they reach the second-best results. They often adopt the framework of meta-learning to update the embeddings of new entities with their support triples. However, when the support set is inaccurate and shares different distributions with the query set, the performance of few-shot methods will be affected. Therefore, T-GEN introduces a stochastic embedding layer to account for the uncertainty, which improves the reliability of its predictions. In our method, we not only adopt the framework of the neural process to quantify the uncertainty but also extract the relational motifs to inject the inductive ability into our model, which outperforms all baseline models. Compared with the 3-shot results, the improvement on the 1-shot is relatively small. The possible reason is that the model is less certain about its predictions given a single observation, which impairs the predictions. Detail studies about uncertainty captured by \ourmethod can be found in Section~\ref{sec:uncertainty}.

In real-world settings, unseen entities emerge simultaneously. Therefore, we also consider the prediction of links between two unseen entities, i.e., Unseen-to-unseen link prediction. We illustrate the performance of our model \ourmethod in Table~\ref{tab:unseen}, where it achieves comparable results to existing state-of-the-art methods. This demonstrates that \ourmethod is capable of inferring hidden relationships among unseen entities and confirms its inductive ability. The I-GEN model ignores the relations between unseen entities, resulting in poor performance.

\begin{table*}[t]
  \centering
  \caption{The results of 1-shot and 3-shot link prediction on FB15k-237 and NELL-995. The best results are highlighted in bold.}
  \label{tab:main-table}
  \resizebox{0.8\textwidth}{!}{%
    \begin{tabular}{p{2cm}cccccccccccccccc}
      \toprule
                               & \multicolumn{8}{c}{\textbf{FB15k-237}} & \multicolumn{8}{c}{\textbf{NELL-995}}   \\ \cmidrule(r){2-9} \cmidrule(r){10-17}
                               & \multicolumn{2}{c}{\textbf{MRR}}        & \multicolumn{2}{c}{\textbf{Hit@1}}    & \multicolumn{2}{c}{\textbf{Hit@3}} & \multicolumn{2}{c}{\textbf{Hit@10}} & \multicolumn{2}{c}{\textbf{MRR}} & \multicolumn{2}{c}{\textbf{Hit@1}} & \multicolumn{2}{c}{\textbf{Hit@3}} & \multicolumn{2}{c}{\textbf{Hit@10}}  \\
      \multirow{-3}{*}{\textbf{Model}} & 1-S                            & 3-S                          & 1-S                       & 3-S                         & 1-S                     & 3-S                       & 1-S                       & 3-S                        & 1-S   & 3-S   & 1-S   & 3-S   & 1-S   & 3-S   & 1-S   & 3-S   \\ \midrule
      TransE                   & .071                          & .120                        & .023                     & .057                       & .086                   & .137                     & .159                     & .238                      & .071 & .118 & .037 & .061 & .079 & .132 & .129 & .223 \\
      DistMult                 & .059                          & .094                        & .034                     & .053                       & .064                   & .101                     & .103                     & .172                      & .075 & .134 & .045 & .083 & .083 & .143 & .131 & .233 \\
      ComplEx                  & .062                          & .104                        & .037                     & .058                       & .067                   & .114                     & .110                     & .188                      & .069 & .124 & .045 & .077 & .071 & .134 & .117 & .213 \\
      RotatE                   & .063                          & .115                        & .039                     & .069                       & .071                   & .131                     & .105                     & .200                      & .054 & .112 & .028 & .060 & .064 & .131 & .104 & .209 \\ \midrule
      R-GCN                    & .099                          & .140                        & .056                     & .082                       & .104                   & .154                     & .181                     & .255                      & .112 & .199 & .074 & .141 & .119 & .219 & .184 & .307 \\
      MEAN                     & .105                          & .114                        & .052                     & .058                       & .109                   & .119                     & .207                     & .217                      & .158 & .180 & .107 & .124 & .173 & .189 & .263 & .296 \\
      LAN                      & .112                          & .112                        & .057                     & .055                       & .118                   & .119                     & .214                     & .218                      & .159 & .172 & .111 & .116 & .172 & .181 & .255 & .286 \\ \midrule
      GMatching                & .224                          & .238                        & .157                     & .168                       & .249                   & .263                     & .352                     & .372                      & .120 & .139 & .074 & .092 & .136 & .151 & .215 & .235 \\
      MetaR                    & .294                          & .316                        & .223                     & .235                       & .318                   & .341                     & .441                     & .492                      & .177 & .213 & .104 & .145 & .217 & .247 & .315 & .352 \\
      FSRL                     & .255                          & .259                        & .187                     & .186                       & .279                   & .281                     & .391                     & .404                      & .130 & .161 & .075 & .106 & .145 & .181 & .253 & .275 \\
      I-GEN	&.348 	&.367 	&.270 	&.281 	&.382 	&.407 	&.504 	&.537 	&.278 	&.285 	&.206 	&.214 	&.313 	&.322 	&.416 	&.426  \\
      T-GEN	&.367 	&.382 	&.282 	&.289 	&.410 	&.430 	&.530 	&.565 	&.282 	&.291 	&.209 	&.217 	&\textbf{.320} 	&.333 	&\textbf{.421} 	&.433  \\ \midrule
      \ourmethod               & \textbf{.371}                 & \textbf{.409}               & \textbf{.289}            & \textbf{.323}              & \textbf{.411}          & \textbf{.453}            & \textbf{.532}            & \textbf{.575}             & \textbf{.283}  & \textbf{.314} & \textbf{.210} & \textbf{.243} & .316 & \textbf{.352} & .419 & \textbf{.452} 
 \\ \bottomrule
    \end{tabular}%
  }
\end{table*}

\begin{table*}[tp]
  \centering
  \caption{The seen-to-unseen and unseen-to-unseen results of 1-shot and 3-shot link prediction on FB15k-237. Bold numbers denote the best results.}
  \label{tab:unseen}
  \resizebox{0.5\textwidth}{!}{%
    \begin{tabular}{lcccccccc}
      \toprule
      & \multicolumn{4}{c}{\textbf{Seen-to-unseen}} & \multicolumn{4}{c}{\textbf{Unseen-to-unseen}} \\ \cmidrule(r){2-5} \cmidrule(r){6-9}
      & \multicolumn{2}{c}{\textbf{MRR}}  & \multicolumn{2}{c}{\textbf{Hit@10}} & \multicolumn{2}{c}{\textbf{MRR}} & \multicolumn{2}{c}{\textbf{Hit@10}} \\
      \multirow{-3}{*}{\textbf{Model}} & 1-S   & 3-S   & 1-S   & 3-S   & 1-S   & 3-S   & 1-S   & 3-S \\ \midrule
        I-GEN & .371	&.391	&.537	&.571	&.000 	&.000 	&.000 	&.000   \\
        T-GEN & .379 	&.396 	&\textbf{.550} 	&.588 	&.185 	&.175 	&.220 	&.201   \\
        \midrule
        \ourmethod & \textbf{.383} 	&\textbf{.422} 	&.549 	&\textbf{.601} 	&\textbf{.204} 	&\textbf{.198} 	&\textbf{.221} 	&\textbf{.220} \\
      \bottomrule
\end{tabular}%
  }
\end{table*}

\subsection{Ablation Study}
\begin{figure}[t]
    \centering
    \includegraphics[width=.5\columnwidth]{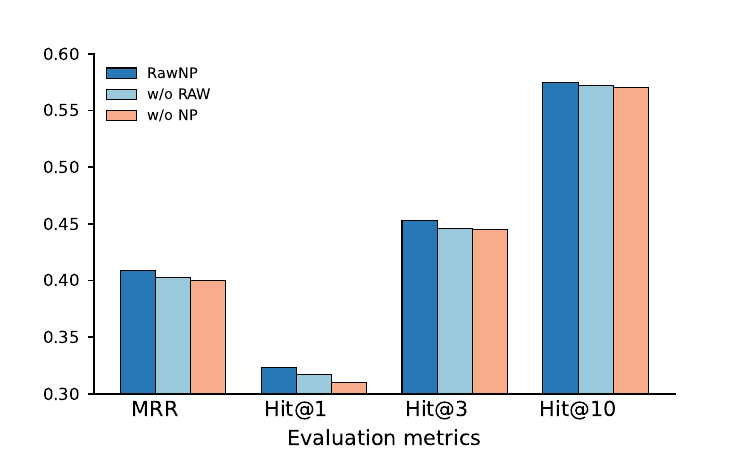}
    \caption{Ablation study on the FB15k-237 dataset.}
    \label{fig:ablation_sdudy}
 \end{figure}

To evaluate the effectiveness of relational anonymous walks (RAW) and neural process (NP), we perform an ablation study by removing each component. The experiment is conducted on the FB15k-237 dataset with a 3-shot support set, and the results are shown in Fig.~\ref{fig:ablation_sdudy}. From the results, we can see that all components (i.e., RAW and NP) are helpful for improving the performance. By removing the RAW, the model ignores the inductive semantic patterns bought by relational motifs, which impairs the ability of inductive reasoning. Without the NP, the model just obtains a deterministic function for the unseen entity, instead of modeling the function distribution. Therefore, the model could suffer from the overfitting problem and fail to generalize to more situations.

\begin{figure}[t]
\centering
\begin{minipage}[t]{0.45\linewidth}
\centering
\includegraphics[width=\textwidth]{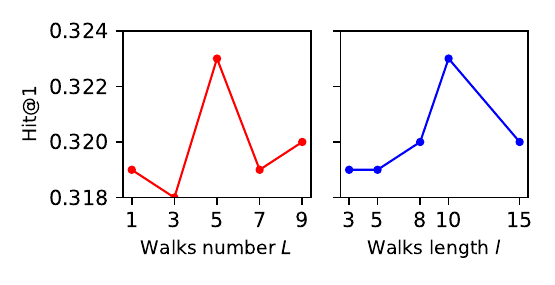}
\caption{Parameter studies on walks number $L$ and walks length $l$.}
\label{fig:exp_1}
\end{minipage}
\hfill
\begin{minipage}[t]{0.45\linewidth}
\centering
\includegraphics[width=\textwidth]{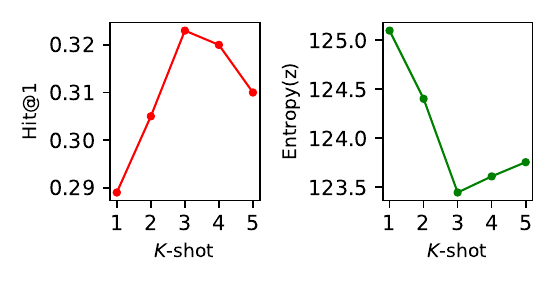}
\caption{Uncertainty analysis under different $K$-shot support set.}
\label{fig:exp_2}
\end{minipage}
\end{figure}

\subsection{Parameters Analysis}
We study the impact of walks number $L$ and walks length $l$ in relational anonymous walks. The results are illustrated in Fig.~\ref{fig:exp_1}. From the results, we can see that the performance of \ourmethod improves as the walks number $L$ increases. The possible reason is that by increasing the walks number, the model could capture more diverse relational motifs, and generate more representative patterns easily. Nevertheless, too many walks could also extract many general patterns that are not dedicated to the unseen entity.
The performance of \ourmethod first increases and then decreases as the walks length $l$ reaches 10. When $l$ is small, the path is too short to represent meaningful patterns (e.g., 1-2-3-4). However, an over large path length could contain redundant motifs that are also not helpful.


\subsection{Uncertainty Analysis}\label{sec:uncertainty}
The major advantage of \ourmethod is able to estimate the uncertainty in its predictions. By using the neural process, we can obtain distribution of the prediction function given the support set. The uncertainty of the model can be evaluated by the entropy of $z$ \cite{naderiparizi2020uncertainty}. The higher the entropy, the more uncertain the model is. We illustrate the Hit@1 under different $K$-shot support sets and calculate the corresponding $Entropy(z)$ by using \cite{singh2016analysis}, which are illustrated in Fig.~\ref{fig:exp_2}.


From the results, we can see that with $K$ increasing, the performance of \ourmethod first improves. This indicates that \ourmethod could adaptively incorporate new observations to enhance the distribution. Then entropy of $z$ also supports the claim. With more data in the support set, the $Entropy(z)$ decreases, meaning the model is more certain about its predictions. The performance of the model slightly decreases when $K \geq 4$, which could be caused by the noise in the support set. The $Entropy(z)$ follows the same trend as the model performance. When the model is more uncertain (i.e., larger entropy), the performance is also worse, which indicates that \ourmethod enables estimating the uncertainty accurately.

\begin{figure}[t]
    \centering
    \includegraphics[width=.48\columnwidth]{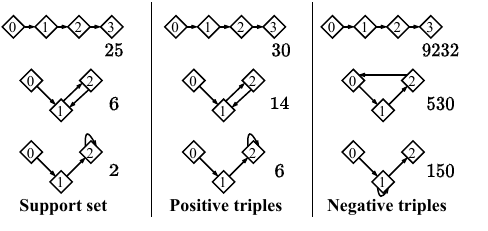}
    \caption{The Top-3 relational motifs and corresponding occurrence numbers extracted for entity $4192$ in FB15k-237.}
    \label{fig:exp_3}
    \vspace{-4mm}
 \end{figure}

\subsection{Case Study of Relational Motif}
In this section, we conduct a case study to illustrate the relational motifs captured by \ourmethod. We first select an unseen entity from FB15k-237, and we illustrate the Top-3 distinctive motifs extracted from its support set, positive triples and negative triples, respectively in Fig.~\ref{fig:exp_3}. From the results, we can see that the relational motifs capture some general semantic patterns (e.g., 1-2-3-4), which widely exist in all sets. However, we also easily find that the motifs from the positive triples are more similar to the motifs from the support set, whereas the motifs from the negative samples cannot match the motifs from the support set. This indicates that \ourmethod could capture the distinguishable relational motifs of the unseen entity for few-shot inductive link prediction. More detailed cases of motif extraction can be found in the \emph{Appendix}.
\section{Conclusion}

In this paper, we propose a novel relational anonymous walk-guided neural process approach for few-shot inductive link prediction on knowledge graphs, named \ourmethod. We first propose a neural process-based approach, which models the distribution over functions conditioned on few-shot observations. 
Then, we propose a novel relational anonymous walk to extract relational motifs to capture general semantic patterns. The comparison against other baseline models demonstrates the superiority of our method. We plan to unify large language models (LLMs) and knowledge graphs to improve the link prediction performance \cite{pan2023unifying}.
\section{Acknowledgement}
This research is supported by NSF of China (No: 61973162), NSF of Jiangsu Province (No: BZ2021013), NSF for Distinguished Young Scholar of Jiangsu Province (No: BK20220080), the Fundamental Research Funds for the Central Universities (Nos: 30920032202, 30921013114), CAAI-Huawei MindSpore Open Fund, and “111” Program (No: B13022).

\clearpage
\bibliographystyle{splncs04}
\bibliography{main}

\clearpage


\appendix
\setcounter{table}{0}
\setcounter{figure}{0}
\setcounter{equation}{0}
\setcounter{algocf}{0}

\section*{Appendix for RawNP: Few-shot Inductive Link Prediction on Knowledge Graphs}
\section{Derivation of ELBO Loss}
Given an unseen entity $u$ and its context data (support set) $\mathcal{C}_u$, the objective of \ourmethod is to infer the distribution $P(z|\mathcal{C}_u)$ from the context data that minimizes the prediction loss on the target data (query set) $\mathcal{D}_u$. The training objective of the neural process can be written as
\begin{equation}
    P(z, e_q|u, r_q, \mathcal{C}_u) = P(z|\mathcal{C}_u) \prod_{\{(u,r_q,?)\in\mathcal{D}_u\}} P(e_q|f_u(u,r_q,z)), \label{eq:obj}
\end{equation}
where $f_u(u,r_q,z)$ denotes the inductive neural process link predictor, and $(u,r_q,?)$ denotes the query to be predicted in the target data.

Following Eq.~\eqref{eq:obj}, the prediction likelihood on target data $\log P\left(e_q|u, r_q, \mathcal{C}_u\right)$ can be written as
\begin{align}
    \log P\left(e_q|u, r_q, \mathcal{C}_u\right)&=\log \frac{P\left(e_q,z|u, r_q,\mathcal{C}_u\right)}{P\left(z|\mathcal{C}_u\right)}, \\
    &=\log P\left(e_q,z|u, r_q,\mathcal{C}_u\right) - \log P\left(z|\mathcal{C}_u\right).
    \label{eq:log}
\end{align}

Assuming that $Q(z)$ is the true distribution of $z$, we can rewrite the Eq.~\eqref{eq:log} as
\begin{align}
    \log P\left(e_q|u, r_q, \mathcal{C}_u\right)=\log \frac{P\left(e_q,z|u, r_q,\mathcal{C}_u\right)}{Q(z)} - \log \frac{P\left(z|\mathcal{C}_u\right)}{Q(z)}.
\end{align}

We integrate both sides with $Q(z)$.
\begin{align}
    \nonumber
    &\int_z Q(z) \log P\left(e_q|u, r_q, \mathcal{C}_u\right) \\
    &=\int_z Q(z) \log \frac{P\left(e_q,z|u, r_q,\mathcal{C}_u\right)}{Q(z)} - \int_z Q(z) \log \frac{P\left(z|\mathcal{C}_u\right)}{Q(z)}.
\end{align}

And then we can rewrite the $\log P\left(e_q|u, r_q, \mathcal{C}_u\right)$ as
\begin{align}
    \nonumber
    &\log P\left(e_q|u, r_q, \mathcal{C}_u\right) \\
    \label{eq:kl}
    &=\int_z Q(z) \log \frac{P\left(e_q,z|u, r_q,\mathcal{C}_u\right)}{Q(z)} + K L\left(Q(z) \| P\left(z|\mathcal{C}_u\right)\right).
\end{align}

Since $K L\left(Q(z) \| P\left(z|\mathcal{C}_u\right)\right) \ge 0$, we can write Eq.~\eqref{eq:kl} as
\begin{align}
    &\log P\left(e_q|u, r_q, \mathcal{C}_u\right) \ge \int_z Q(z) \log \frac{P\left(e_q,z|u, r_q,\mathcal{C}_u\right)}{Q(z)}, \\
    &=\mathbb{E}_{Q(z)} \log \frac{P\left(e_q,z|u, r_q, \mathcal{C}_u\right)}{Q(z)}, \\
    &=\mathbb{E}_{Q(z)}\left[\log P\left(e_q|u, r_q, z\right)+\log \frac{P\left(z|\mathcal{C}_u\right)}{Q(z)}\right], \\ 
    &=\mathbb{E}_{Q(z)}\left[\log P\left(e_q|u, r_q, z\right)\right]-K L\left(Q(z) \| P\left(z|\mathcal{C}_u\right)\right),\label{equ:elbo_loss}
\end{align}
where $Q(z)$ represents the true posterior distribution of $z$, which is intractable. To address this problem, we approximate it with $Q\left(z|\mathcal{C}_u, \mathcal{D}_u\right)$ calculated by the encoder aggregating data from both $\mathcal{C}_u$ and $\mathcal{D}_u$ during training. In Eq.~\eqref{equ:elbo_loss}, the first term is to improve the prediction accuracy by maximizing the prediction likelihood. By minimizing the KL divergence in the second term, we encourage the encoder to infer the target posterior distribution  with limited context data.  

\section{Datasets}
The statistics of two benchmark datasets are shown in Table~\ref{tab:dataset}.

\begin{table}[th]
  \caption{Statistics of two benchmark datasets.}
  \label{tab:dataset}
  \centering
  \resizebox{.6\textwidth}{!}{%
    \begin{tabular}{@{}ccccccc@{}}
      \toprule
      Dataset   & Entities & Relations & Triples & Training & Validation & Test \\ \midrule
      FB15k-237 & 14,514   & 237       & 310,116 & 2,500   & 1,000   & 1,500  \\
      NELL-995  & 75,492   & 200       & 154,213 & 1,500   & 600     & 900  \\ \bottomrule
    \end{tabular}%
  }
\end{table}

\section{Algorithms of Training and Testing Process}
\begin{algorithm2e}[t]
    \caption{The training process of \ourmethod}\label{alg:train}
    \KwIn {Knowledge graph $\mathcal{G}$; Training entities $\mathcal{E}_{train}$}
    \KwOut {Model parameters $\Theta$}
    \While{not done}{
        Sample an unseen entity $
        \mathcal{T}_u = \{\mathcal{C}_u, \mathcal{D}_u\}$ from $\mathcal{E}_{train}$\;
        Generate relational motif representation $M'_e$ using RAW\;
        Generate unseen entity representation $u^{\prime}$ using I-RGNN\;
        Generate prior distribution $P(z|\mathcal{C}_u)$ using RAW-guided neural process encoder\;
        Generate the variational posterior distribution $Q\left(z|\mathcal{C}_u, \mathcal{D}_u\right)$ by feeding $\mathcal{C}_u$, $\mathcal{D}_u$ into the encoder\;
        Sample a $z$ from the posterior distribution $Q\left(z|\mathcal{C}_u, \mathcal{D}_u\right)$\;
        Optimize $\Theta$ using ELBO loss\;
    }
\end{algorithm2e}
\textbf{Training.} We illustrate the training process shown in Algorithm \ref{alg:train}. In the training phase, we first sample an unseen entity $u$ together with its support set $\mathcal{C}_u$ and query set $\mathcal{D}_u$ (Line 2). Then, we generate the relational motif representation $M'_e$ for each entity $e$ in the support set using relational anonymous walk (RAW) (Line 3). Then, we generate the representation of unseen entity $u'$ using I-RGNN (Line 4). Combining the representations of each triple in support set $\mathcal{C}_u$, we define the prior distribution $P(z|\mathcal{C}_u)$ using a RAW-guided neural process encoder (Line 5). Since the true distribution of $Q(z)$ is intractable, in the training stage, we feed both $\mathcal{C}_u$ and $\mathcal{D}_u$ together into the encoder to generate the variational posterior distribution $Q\left(z|\mathcal{C}_u, \mathcal{D}_u\right)$ (Line 6) to approximate the true distribution of $Q(z)$. Last, we sample a $z$ from the posterior distribution to realize the prediction function (Line 7). By optimizing the ELBO loss, we can not only maximize the prediction likelihood but also encourage the encoder to infer the distribution with limited context data (Line 8).

\vspace{2mm}
\noindent \textbf{Testing.} In the testing stage, given an unseen entity $u$, we first generate the unseen entity representation $u^{\prime}$ and the prior distribution $P(z|\mathcal{C}_u)$ from its support set. Since we only have the support set during the testing phase, we sample a $z$ from the prior distribution $P(z|\mathcal{C}_u)$. Then, we obtain the representation of possible entity $e'_q$ and its corresponding relational motif representation $M'_{e_q}$. Finally, we adopt the inductive neural process link predictor to predict the possible entity $e_q$ for a query $q=\left(u, r_q, ?\right)$.

\section{Baseline Models}
We select a series of following baseline models for comparison, which can be divided into three categories: (1) \textbf{Traditional KGC methods}, including TransE \cite{bordes2013translating}, DistMult \cite{yang2014embedding}, ComplEx \cite{trouillon2016complex}, RotatE \cite{sun2019rotate}; (2) \textbf{GNN-based methods}, including R-GCN \cite{schlichtkrull2018modeling}, MEAN \cite{hamaguchi2017knowledge}, LAN \cite{wang2019logic}; (3) \textbf{Few-shot inductive methods}, including GMatching \cite{xiong2018one}, MetaR \cite{chen2019meta}, FSRL \cite{zhang2020few}, GEN \cite{baek2020learning}. We describe these baseline models in detail as follows:

\vspace{2mm}
\noindent \textbf{Traditional KGC methods.} This group of methods contains the translation model and the semantic matching model. The translation model focuses on the use of relationships between entities and the semantic matching model adopts semantic similarity to mine the potential semantics. They are all transductive methods.
\begin{itemize}
    \item TransE \cite{bordes2013translating} is a typical translation embedding model, which represents both entities and relations as vectors in the same space.
    \item RotatE \cite{sun2019rotate} extends the TransE with a complex operation, which defines each relationship as the rotation from the source entity to the target entity in the complex vector space.
    \item DistMult \cite{yang2014embedding} represents the relationship between the head and tail entity in a bi-linear formulation to capture the semantic similarity.
    \item ComplEx \cite{trouillon2016complex} introduces embeddings on a complex space to handle asymmetric relations.
\end{itemize}

\vspace{2mm}
\noindent \textbf{GNN-based methods.} Thanks to the inductive ability of GNNs, this group of methods utilizes the graph structure to model relational data and predict links inductively.
\begin{itemize}
    \item R-GCN \cite{schlichtkrull2018modeling} extends the graph convolutional network to model multi-relational data.
    \item MEAN \cite{hamaguchi2017knowledge} utilizes a GNN-based neighboring aggregation scheme to generate the embedding of entities.
    \item LAN \cite{wang2019logic} further applies the attention mechanisms to consider relations with neighboring information by extending the MEAN model.
\end{itemize}

\vspace{2mm}
\noindent \textbf{Few-shot inductive methods.} This group of methods is all under the meta-learning framework, which can predict links for an unseen entity with few-shot related triples. 
\begin{itemize}
    \item GMatching \cite{xiong2018one} introduces a local neighbor encoder to learn entity embeddings and an LSTM matching network to calculate the similarity.
    \item MetaR \cite{chen2019meta} adapts to unseen relations by a relation-meta learner and updates the parameter under the meta-learning framework.
    \item FSRL \cite{zhang2020few} utilizes an LSTM encoder to summarize the support set information.
    \item GEN \cite{baek2020learning} includes two variants, i.e., the inductive version I-GEN and the transductive version T-GEN. I-GEN embeds an unseen entity to infer the missing links between seen-to-unseen entities while T-GEN adopts an additional transductive stochastic layer for unseen-to-unseen link prediction.
\end{itemize}

\section{More Detailed Cases of Motif Extraction}
\begin{figure}[ht]
    \centering
    \includegraphics[]{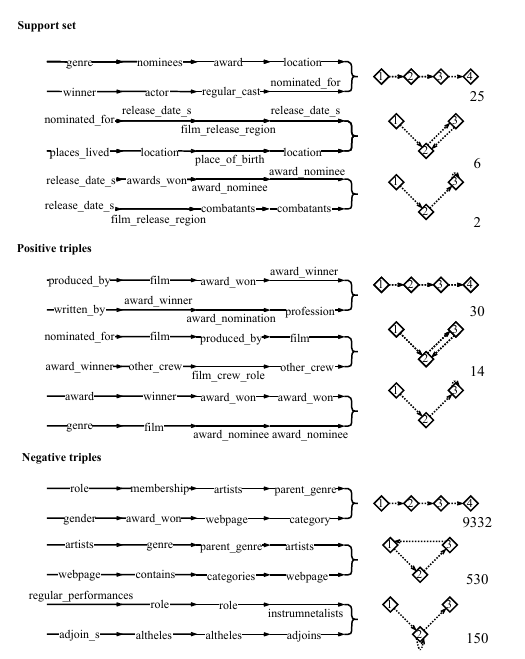}
    \caption{Realistic relations for relational motifs in entity 4192}
    \label{fig:entity_4192}
\end{figure}

\begin{figure}[ht]
    \centering
    \includegraphics[]{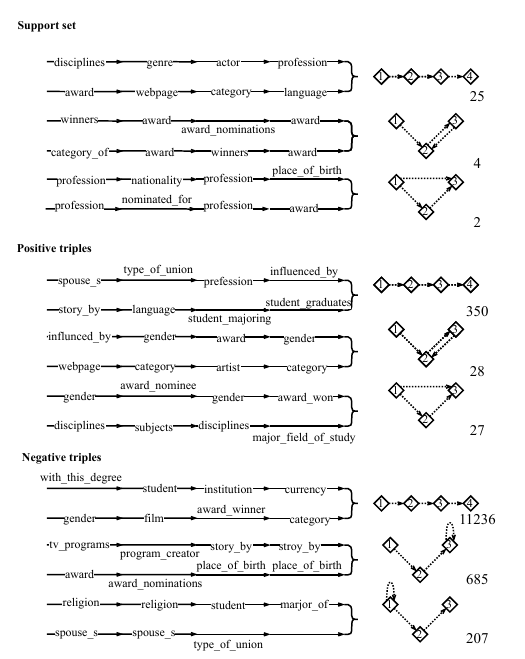}
    \caption{Realistic relations for relational motifs in entity 3953}
    \label{fig:entity_3953}
\end{figure}

\begin{figure}[ht]
    \centering
    \includegraphics[]{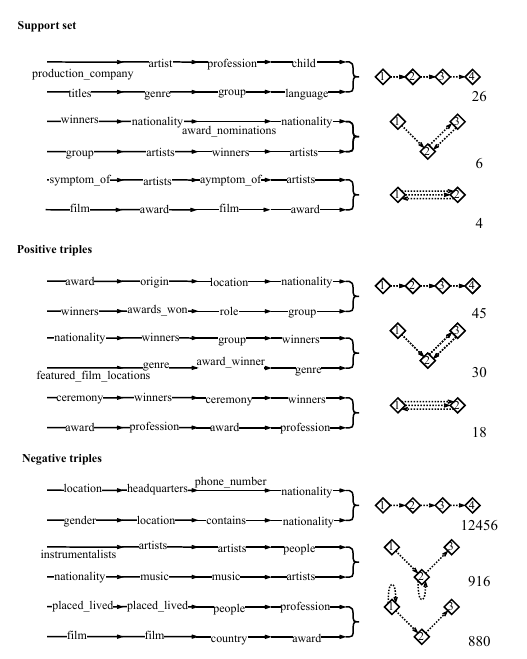}
    \caption{Realistic relations for relational motifs in entity 7276}
    \label{fig:entity_7276}
\end{figure}

In this section, we illustrate three detailed examples of motif extraction for unseen entities in Fig.~\ref{fig:entity_4192}, \ref{fig:entity_3953}, and \ref{fig:entity_7276}, respectively. In each figure, we first illustrate the Top-3 distinctive motifs extracted by relational anonymous walk. For each motif, we illustrate two relational paths that can be anonymized to the same motif. 

From the results, we can find that the relational paths in support sets, positive triples, and negative triples are quite distinct. Therefore, we cannot predict triples by matching the relational paths with ones from the support set. This is consistent with our motivation that relational paths are not general enough for inductive link prediction on knowledge graphs. On the other hand, the different relational paths can be mapped into a general motif. For example, as shown in Fig.~\ref{fig:entity_4192}, paths $award\to winner\to award\_won\to award\_won$ and $genre\to film\to award\_nomince\to award\_noiminee$ can be mapped into the same motif $1\to 2\to 3\to 2$. This motif can also be found in the support set that can be mapped by two other distinct paths, e.g., $relase\_data\_s\to awards\_won\to award\_nominee\to award\_nominee$ and $release\_data\_s\to file\_release\_region\to combatants\to combatants$. In this way, by extracting the relational motifs, we can find the general semantic patterns that are more suitable for inductive link prediction on knowledge graphs.


\end{document}